\documentclass{article}
\usepackage{iclr2027_conference,times}

\usepackage{amsmath,amsfonts,bm}

\def\eqref#1{equation~\ref{#1}}

\def\1{\bm{1}}

\DeclareMathAlphabet{\mathsfit}{\encodingdefault}{\sfdefault}{m}{sl}
\SetMathAlphabet{\mathsfit}{bold}{\encodingdefault}{\sfdefault}{bx}{n}

\usepackage{hyperref,url,graphicx,booktabs,amsmath,array,longtable}
\hypersetup{colorlinks=true,linkcolor=black,citecolor=black,urlcolor=black,
 pdftitle={When Local Definitions Are Missing: Stable Errors in Language Models},
 pdfauthor={Zengqing Wu and Chuan Xiao}}
\title{When Local Definitions Are Missing:\\ Stable Errors in Language Models}
\author{Zengqing Wu$^1$ \quad Chuan Xiao$^{1,2}$\thanks{Corresponding author: \href{mailto:chuanx@nagoya-u.jp}{\texttt{chuanx@nagoya-u.jp}}. Our source code is available at \url{https://github.com/wuzengqing001225/llm-local-meaning-diagnostics}.}\\
{\normalfont $^1$University of Osaka \quad $^2$Nagoya University}}
\iclrfinalcopy
\begin{document}
\maketitle
\lhead{Preprint}
\begin{abstract}
Contracts, technical manuals, and online communities often give familiar words special meanings and define them in one place. A language model that reads a passage without that definition may use the usual meaning and give an answer that sounds right but is wrong. We study this with multiple-choice questions whose correct answers depend on a local definition. The questions come from a synthetic vocabulary, commercial contracts, Reddit community glossaries, and the documentation of a decentralized exchange. We ask three language models each question with and without the definition. Without using the answer key, we then test whether their repeated answers and stated confidence reveal their own errors. Separately, we give a small open model the answer key and test whether it can predict which questions larger models get wrong. We report three findings. (1) Familiar words that a source redefines cause the most errors. Models often repeat the same wrong answer, and neither repeated answers nor stated confidence reliably flag it. (2) The source definition fixes most errors on synthetic questions, but a definition drawn from the model's own prior knowledge does not. In real contracts, the source definition fixes only about half of the errors. (3) Given an answer key, a small model can rank questions by how likely larger models are to get them wrong. These results call for tests that check answers against the full source, including its local definitions.
\end{abstract}

\section{Introduction}
\label{sec:intro}
Language models are now used to answer questions about contracts, technical manuals, and online discussions. These documents often give familiar words special meanings. A contract defines its capitalized terms. A protocol manual may give a word like pool or position a precise technical scope. Online communities keep glossaries of their own jargon \citep{lucy-bamman-glossaries}. In many game communities, for example, a tank is a player who takes hits to protect the group, not a vehicle or a container.

Consider a distribution agreement in CUAD \citep{cuad} that uses Trademarks to mean only the names in Exhibit~A. Another clause requires the distributor to report unauthorized use of the Trademarks. Must the distributor report a similar name that is not on the list? In our contract tests, a language model gets this question wrong without the definition and answers correctly once it sees the definition. The reporting clause alone gives no sign that the term has a narrow meaning. This matters because systems often show the model only a few retrieved passages. In our CUAD retrieval experiments, the relevant definition appears in only 29\% to 44\% of the retrieved passage sets. An answer can fit the passages the model sees and still be wrong under the contract as a whole.

Why does this happen? Speakers and writers design what they say for the people they expect to understand it. They rely on the knowledge they believe they share with those people, which is their common ground, and members of a community can take what the community knows as common ground \citep{clark1982audience}. The reporting clause is written for a reader who already knows the definitions, so the writer does not restate them. A language model is a new kind of reader. Its common ground with the writer is largely general usage, so a local meaning may lie outside it. What the model must be told depends on the gap between its prior knowledge and the document.

A reader needs two abilities here: following a definition that it is given, and recognizing that a definition is missing. Most knowledge-conflict studies give a model competing evidence and check which source it follows. We test what happens when the definition is missing and whether the model's answers give any warning.

Our experiments use four sets of multiple-choice questions. Each question has four options and an answer key that a model drafted from the source. The first set is synthetic. It uses invented vocabularies for a trading desk and a restaurant group, and we fix each term's class before generating any text. The classes are common words that we redefine, invented words, and words used in their usual meaning. The other three sets come from CUAD contracts, from the glossaries of 35 Reddit communities, and from the documentation of a decentralized exchange. We treat the last two as exploratory because their sources often leave more than one possible answer. We give or withhold the definition directly. Retrieval appears only in the motivating estimate above and in appendix experiments.

Three language models, GPT-5.6-terra, GPT-6-astra, and DeepSeek Flash (V4/4.1), act as readers. To check that readers can follow a definition when it is given, we use 1,244 human-written redefinition items from Inverse Scaling \citep{inverse-scaling}, in which the prompt states a new meaning for a word. GPT-5.6-terra and GPT-6-astra follow it on 95.0\% and 99.8\% of the items, and DeepSeek V4.1 Flash on 63.0\%. For the GPT readers, applying a given definition is rarely the problem. On our diagnostic questions, each reader answers without the definition and again with definition text. To test whether a reader's answers warn of its errors, we draw ten answers per question, record stated confidence, and also test TypeSafe AI's Jev, a model that reports its own confidence \citep{jev}. To predict which questions are hard, we use small open models, Qwen2.5 with 0.5B to 14B parameters and Llama-3.1-8B, that read each question without the definition. We report three findings:
\begin{enumerate}
\item \textbf{Redefined common words produce stable errors, and confidence signals do not reliably flag them.} These words have the highest error rate among the synthetic classes. For DeepSeek V4 Flash it is 44\%, against 15\% for words used in their usual meaning. We call an error stable when asking again gives the same wrong answer. Across ten resampled answers, 41\% of DeepSeek V4 Flash's errors on the synthetic questions repeat in every sample, and majority voting leaves 65\% of the errors on redefined common words in place. Agreement across samples and stated confidence rank these errors barely better than chance, with AUROC 0.56 and 0.46 where 0.5 is chance (Sections~\ref{sec:stable} and~\ref{sec:detect}).
\item \textbf{Source definitions provide information that prior knowledge lacks.} On the synthetic questions, the source definition raises DeepSeek V4 Flash from 70\% to 97\% accuracy. A gloss that a model drafts from the source does nearly as well (95\%), but a gloss that the reader writes from its own prior knowledge raises it only to 71\% (Section~\ref{sec:repair}).
\item \textbf{A small model can help select difficult evaluation questions.} Given the answer key, the 7B-parameter Qwen2.5 model ranks the synthetic questions that DeepSeek V4 Flash gets wrong with an AUROC of 0.85. The ranking improves with model size, from 0.70 with the 0.5B model to 0.87 with the 14B model. The score also ranks errors on usual-meaning words, so it reflects general difficulty and not only missing definitions (Section~\ref{sec:predict}).
\end{enumerate}

In short, models largely follow a definition stated in the prompt, but when the definition is missing, the confidence signals we tested give little warning. The definition should come from the source. Evaluations should check answers against the full source and not only against the passage the model saw. Linking diagnostic questions to the local definitions in a source gives one way to do this for a document collection.
\section{Related work}
\label{sec:related}
\paragraph{Error types and missing evidence.}
Hallucination and knowledge-conflict research separates errors by the standard that an answer is judged against. A factuality error disagrees with world knowledge, and a faithfulness error disagrees with the supplied input \citep{space}. In a knowledge conflict, the input contradicts what the model has learned, and the question is which source the model follows \citep{longpre-etal-2021-entity,kc-survey,conflictbank,knowledge-conflict-faithfulness}. Decoding and steering methods aim to make models rely more on the input \citep{cad,adacad,coiecd,contextfocus,taming-kc}. Table~\ref{tab:error-types} places our case among these categories. A missing local definition differs because the needed evidence is absent from the input, and the answer is correct or wrong only with respect to the full source document.
\begin{table}[t]
\centering\small
\caption{Error types by the standard an answer is judged against. The first three categories are used in hallucination and knowledge-conflict research \citep{space,longpre-etal-2021-entity,kc-survey}.}
\label{tab:error-types}
\begin{tabular}{@{}lll@{}}
\toprule
Error type & Answer judged against & Needed evidence in the input\\\midrule
Factuality error & World knowledge & Not required\\
Faithfulness error & Supplied input & Present\\
Knowledge conflict & Input versus learned knowledge & Present, contradicts the prior\\
Missing local definition & Full source document only & Absent\\\bottomrule
\end{tabular}
\end{table}
In one test with Qwen2.5-14B-Instruct on 1,200 CUAD questions, contrastive decoding over the retrieved passages lowers accuracy by 2.8 points $[-4.4,-1.0]$.

Earlier work on missing evidence includes fact-checking tests that remove evidence and ask whether the remaining evidence is sufficient \citep{atanasova-2022-insufficient}. MIGRES uses models to find missing facts for retrieval in multi-hop question answering \citep{wang-2025-migres}. We focus on one kind of missing evidence: a source's definition of a familiar term.

\paragraph{Definitions, local language, and term selection.}
Redefinition tasks ask whether a model follows a stated meaning \citep{inverse-scaling,redefine-pitfalls}. Studies of concept guidelines and label definitions ask how models use instructions \citep{fonseca-2024-guidelines,mohammadi-2025-definitions}. Work on community language, slang, and ordinary meaning in law examines how meaning varies and how models interpret it \citep{lucy-bamman-glossaries,slang,legal-interpretation}. We ask the same question about a term with and without its source definition. We also measure how far the source meaning lies from the model's prior gloss.

Methods that pick out terms needing attention from text alone make an assumption about the reader. Inverse document frequency gives more weight to terms that occur in fewer documents of a collection \citep{sparck2004statistical}. TextRank ranks words by their centrality in a co-occurrence graph \citep{mihalcea2004textrank}. Lexical semantic change detection looks for terms used differently in two corpora \citep{lsc-survey,semeval2020-lsc}. These methods score terms without a model of what the reader already knows. We test term frequency and rarity as ranking scores, and we do not test TextRank or semantic change detectors.

\paragraph{Error detection and question difficulty.}
Sampling consistency, semantic entropy, and stated confidence can help rank errors \citep{selfcheckgpt,semantic-uncertainty,semantic-entropy,verbal-confidence}. We test these signals when a local definition is missing. Our proxy score also belongs to work that estimates shared item difficulty through item response theory or answer plausibility \citep{vania-etal-2021-comparing,byrd-srivastava-2022-predicting,mozafari-etal-2026-question}. SlimPLM uses a small model to decide what to retrieve for an incoming question \citep{tan-etal-2024-small}. We instead use a small model offline to screen questions that already have keys. This use can sit alongside retrieval and long-context systems \citep{rag,lost-middle}.
\section{Setting}
\label{sec:meaning}
\paragraph{Local meanings and diagnostic questions.}
A \emph{local meaning} is the meaning or scope that a document or community gives to a term. The model that answers a question is the \emph{reader}. In the bare condition, the reader does not see the local definition, so its answer shows its default reading. A diagnostic question asks about one use of a term and tests a consequence of its local meaning. Each question has four options and a key drafted by a model from the source. The key is a proposed answer, not an expert judgment.

We use three kinds of definition text. The source definition is the original wording from the document or glossary. A drafted gloss is a short explanation that a model writes from the source. A prior gloss is the reader's own explanation of the term under a domain prompt, written without the source.

An answer can need local information for different reasons. The usual meaning may conflict with the local definition, the local boundary may be unspecified, or the needed fact may lie outside the definition. Table~\ref{tab:cases} describes these situations. We have not assigned them to individual questions.
\begin{table}[t]
\centering\small
\caption{Three possible reasons an answer may need local information. These are interpretive cases, not annotated question categories.}
\label{tab:cases}
\begin{tabular}{@{}>{\raggedright\arraybackslash}p{2.7cm}>{\raggedright\arraybackslash}p{5.4cm}>{\raggedright\arraybackslash}p{5.0cm}@{}}
\toprule
Situation & Missing or misleading information & Possible remedy\\\midrule
Meaning conflict & The usual meaning supports a different answer from the local definition. & Supply the source definition and its scope.\\
Scope unspecified & The general concept fits, but its members or boundary in this document are unknown. & Supply the local boundary or binding.\\
Other information & The needed fact or rule is outside the definition. & Retrieve the relevant evidence.\\\bottomrule
\end{tabular}
\end{table}

\paragraph{Corpora.}
In the synthetic corpus, we design vocabulary for a trading desk and a restaurant group. Its three term classes are redefined common words, invented words, and words used in their usual meaning. CUAD contains 510 commercial contracts \citep{cuad}. Our Reddit data come from 35 communities with local glossaries. The DeFi data come from the documentation of a decentralized exchange protocol. The appendix gives the material used to draft the questions and identifies the definition text used in each test.

The writer of the synthetic questions saw examples of usage without the source definition. For Reddit and DeFi, the writers also saw the source entries. For CUAD, they saw an extracted definition and a passage where the term occurred. These drafting procedures can leave answer cues or produce keys that the source does not support.

Two models separately checked 20 questions per corpus against the preserved sources. We selected these questions using a fixed hash. Both models supported the stored key on 18 synthetic, 13 CUAD, 3 Reddit, and 5 DeFi questions. This was a model screen. The author review covers a selected subset and is described in Appendix~\ref{app:source-screen}.

The controlled synthetic tests and the CUAD diagnostic results carry most of the analysis. We treat Reddit and DeFi as exploratory because the preserved sources often leave more than one possible answer. A separate human-written task tests whether models can follow an explicit redefinition.

\paragraph{Models.}
The readers are GPT-5.6-terra, GPT-6-astra, and DeepSeek Flash (V4/4.1). We used DeepSeek V4 Flash for the synthetic questions and DeepSeek V4.1 Flash for later runs. TypeSafe AI's Jev returns decision probabilities and a confidence score \citep{jev}. As small proxy models, we use Qwen2.5 models from 0.5B to 14B parameters and Llama-3.1-8B \citep{qwen,llama}. A proxy reads a question without the definition and assigns a probability to each answer option. Appendix~\ref{app:models} lists the model identifiers and decoding settings.

\paragraph{Ranking errors.}
Several tests ask how well a score ranks wrong answers above correct ones. We measure this with AUROC, the probability that a randomly chosen wrong answer receives a higher score than a randomly chosen correct answer. A value of 0.5 means chance and 1 means a perfect ranking. To compute confidence intervals, we resample contracts for CUAD, communities for Reddit, and term records otherwise \citep{significance}.
\section{Stable errors on locally defined terms}
\label{sec:stable}
\paragraph{Following versus recognizing a definition.}
We first test whether readers follow a rule when the prompt states it. On 1,244 human-written Inverse Scaling items \citep{inverse-scaling}, GPT-5.6-terra follows the stated meaning on 95.0\% of items and GPT-6-astra on 99.8\%. When we remove the redefinition, both use the usual meaning on about 98\% of items. DeepSeek V4.1 Flash follows the stated meaning on 63.0\%.

We then withhold the definition. Without it, DeepSeek V4 Flash gets 28.5\% of the matched synthetic questions wrong, and GPT-6-astra gets 18.2\% wrong. For GPT-5.6-terra, the weighted error estimate across the 6,529 scored CUAD questions is 13.2\%. Error rates in the exploratory Reddit and DeFi settings range from 14.6\% to 30.1\%. The two tests use different questions, and the error rates describe our diagnostic questions and not live user requests.

\begin{table}[t]\centering\footnotesize\setlength{\tabcolsep}{4pt}
\caption{Examples from the contract validation set. GPT-5.6-terra answers each question incorrectly without the definition and correctly with it. The usual meanings suggest broader readings than the contracts allow. Risk percentiles refer to the 6,529 scored questions.}
\label{tab:examples}
\begin{tabular}{@{}p{1.5cm}p{3.6cm}p{3.9cm}p{2.9cm}r@{}}\toprule
Term & Local definition & Question (abridged) & Answer implied by the definition & Risk\\\midrule
Products & dietary supplements manufactured within the United States & Which products must the endorser have used to give honest findings? & only U.S.-made supplements & 97th\\
Trademarks & the trademarks and trade names identified in Exhibit~A & Must the distributor report a confusable third-party name not in Exhibit~A? & report only names listed in Exhibit~A & 99.7th\\
Term & the period from the Signing Date to the End Date & During which period must the required insurance stay in force? & from signing to the stated end date, not until termination & 98th\\
\bottomrule\end{tabular}\end{table}

\begin{figure}[t]
\centering\includegraphics[width=.85\linewidth]{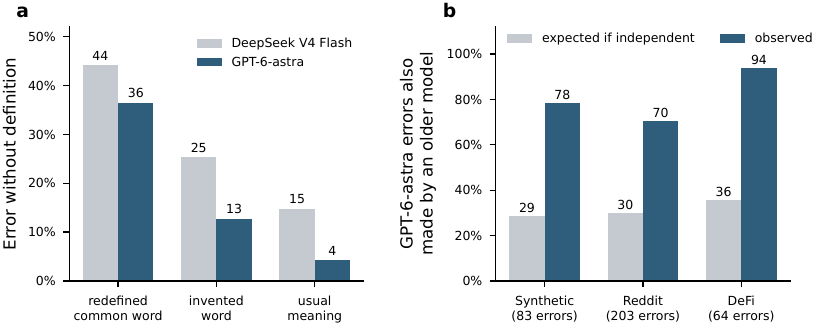}
\caption{(a) Error rates without definitions for the three synthetic term classes. Errors are most common on redefined common words. (b) Share of GPT-6-astra's errors that another reader makes on the same questions. Under independence, the expected share is that reader's overall error rate on those questions. For DeFi, the comparison counts an error by either DeepSeek V4.1 Flash or GPT-5.6-terra.}
\label{fig:phenomenon}
\end{figure}

\paragraph{Designed term classes.}
DeepSeek V4 Flash gets 44.2\% of the matched synthetic questions about redefined common words wrong (Figure~\ref{fig:phenomenon}). Its error rates are 25.3\% for invented words and 14.8\% for words used in their usual meaning. GPT-6-astra shows the same ordering, at 36.5\%, 12.7\%, and 4.2\%. This pattern fits models falling back on a familiar interpretation, and the contract examples show what a broader reading can miss. This ordering does not follow rarity. Invented words are the rarest, but redefined common words cause more errors.

\paragraph{Repeated answers.}
We call an error stable when the reader gives the same wrong answer every time we ask. DeepSeek V4 Flash first gave a greedy answer to each of 761 synthetic questions. It then gave ten sampled answers at temperature 1.0. The greedy answer was wrong on 229 questions. On 93 of them (40.6\%), all ten samples repeated the same wrong answer. A majority vote still gives the wrong answer on 142 questions (62\%) and corrects only 38\% of the initial errors. Voting helps less for redefined common words, where 65\% of the initial errors remain, than for usual-meaning words, where 43\% remain. This test is related to self-consistency \citep{self-consistency}, but we vote on answers without sampling explicit chains of thought.

\paragraph{Overlap across readers.}
The readers often get the same questions wrong. DeepSeek V4 Flash makes 78\% of GPT-6-astra's synthetic errors on the shared questions. On Reddit, the overlap with DeepSeek V4.1 Flash is 70\%. On DeFi, the share is 94\% when we count errors by either DeepSeek V4.1 Flash or GPT-5.6-terra. If the error sets were independent, the expected shares would be 29\%, 30\%, and 36\%. The readers differ in model family and size, so the overlap points to shared difficult questions. It cannot separate the effect of model generation from the role of local meaning.
\section{Confidence signals give little warning}
\label{sec:detect}
\begin{figure}[t]
\centering\includegraphics[width=\linewidth]{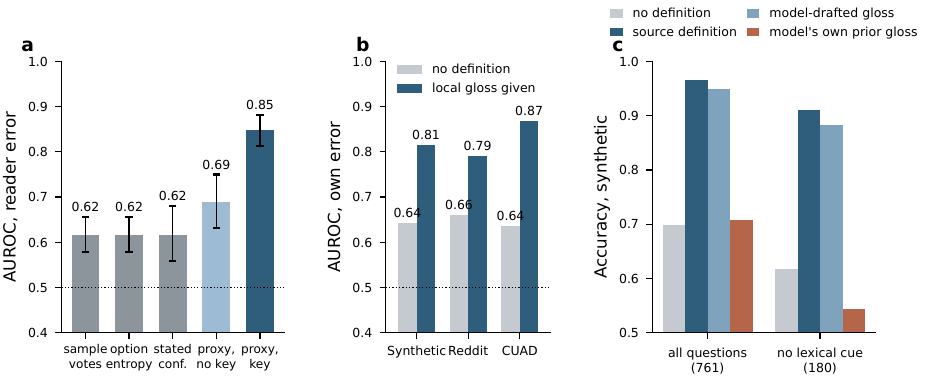}
\caption{(a) Error-ranking AUROC for DeepSeek V4 Flash on 456 synthetic questions. We obtain 95\% intervals by resampling terms. Gray bars show signals from ten sampled answers and stated confidence, which use no key. The light bar shows key-free uncertainty for the 7B proxy. The dark bar shows the proxy score $R$ (Section~\ref{sec:predict}), which uses the answer key. The dotted line marks chance. (b) How well Jev's confidence ranks its own errors, with no definition and with a drafted gloss. (c) DeepSeek V4 Flash accuracy with no definition, the source definition, a drafted gloss, or a prior gloss. On the right, the lexical screen finds no option singled out by word overlap with any of the three notes.}
\label{fig:silent}
\end{figure}

If a model could tell when it lacks a definition, we could supply one only in those cases. This resembles a teacher who rechecks only the answers where a student hesitated. A confident wrong answer is an error that the model does not know it makes, and a hesitation check cannot catch it. We ask whether the reader's own signals separate its wrong answers from its correct ones on the 456 synthetic questions shared with the proxy. The signals are disagreement across ten samples, the entropy of the option votes, and the negative of stated confidence \citep{selfcheckgpt,semantic-uncertainty,verbal-confidence}. None needs an answer key. For DeepSeek V4 Flash, their AUROCs are 0.616 $[0.579,0.656]$, 0.616 $[0.579,0.656]$, and 0.617 $[0.558,0.681]$, respectively (Figure~\ref{fig:silent}). For redefined common words the values are lower: 0.557, 0.557, and 0.463. On invented and usual-meaning words, they range from 0.60 to 0.72. We measure entropy over votes for fixed options, which is narrower than semantic entropy over free-form answers \citep{semantic-entropy}.

We also test the proxy's own uncertainty, taken as one minus its largest option probability, which does not use the key either. On redefined common words, this score reaches only 0.537 and 0.523 for the two readers. In these tests, confidence and repeated answers give little assurance that the definition was unnecessary.

Jev reports a confidence score for its own answers. Without a definition, this score ranks Jev's errors at AUROC 0.643 on synthetic questions, 0.660 on Reddit, and 0.636 on CUAD. With a drafted gloss, the values rise to 0.815, 0.790, and 0.868. Once the definition is given, the same confidence signal flags errors well. The gloss also changes which questions Jev gets wrong, so this comparison does not isolate calibration as the cause.

\section{What source definitions repair}
\label{sec:repair}
Because the reader's own signals give little warning, the missing information has to come from outside. We ask where it should come from. Supplying definitions lowers the error rate in every setting we tested (Table~\ref{tab:prediction}).

\paragraph{Source definitions and prior glosses.}
The writer of the 761 synthetic questions saw usage examples but not the definition. DeepSeek V4 Flash reaches an accuracy of 0.699 without a definition. The source definition raises it to 0.966, and a drafted gloss based on that source reaches 0.949. For the prior gloss, we ask the reader to explain the term under a domain prompt without showing the source. This gloss reaches only 0.708. The source definition fixes 209 errors and turns 6 correct answers into errors. The prior gloss fixes 86 errors but turns 79 correct answers into errors.

We also examine the 180 questions for which the overlap screen flags none of the three definition notes. Accuracy is 0.617 without a note. It rises to 0.911 with the source definition and drops to 0.544 with the prior gloss. The screen checks for one kind of wording overlap, so other answer cues may remain. On these questions, the source text helps in a way the prior gloss does not, and the drafted gloss also gives a large improvement. On these questions, the needed information came from the source and not from the prior gloss.

\paragraph{Contract definitions.}
Giving the reader the extracted contract definitions raises accuracy from 0.787 to 0.840. It fixes 163 errors and makes 83 correct answers wrong, and 157 of the 320 initial errors remain. A definition can refer to other parts of the document. In one reviewed question, for example, the answer depends on an exhibit that was redacted from the contract. Such references may explain part of the remaining errors. Other possible reasons are question defects, incomplete notes, and mistakes in applying the definition, and the model-generated annotations cannot tell us how much each contributes. Our small Reddit and CUAD development comparisons have not shown that selecting definitions by score improves on term frequency.

\paragraph{Wording and scope.}
In the exploratory corpora, the wording of a definition changes which answers it helps. On Reddit glossary questions, the source entry raises accuracy from 0.657 to 0.932. It fixes 243 errors and makes 9 correct answers wrong. On the DeFi source-entry subset, accuracy rises from 0.672 to 0.940, with 70 errors fixed and 3 correct answers made wrong. With drafted glosses, the counts are 171 fixes and 58 harms on Reddit and 42 fixes and 20 harms on DeFi. The prior gloss on Reddit harms more answers than it fixes (117 versus 82). The question writers saw the source entries in both corpora, so these comparisons describe the texts used in the study and do not measure definition quality independently. The appendix reports how $\Delta p$ ranks repairs of observed bare errors on the same questions.

\paragraph{Public task comparisons.}
In CUAD clause extraction, adding definitions changes token F1 by at most 0.3 points. Clause extraction asks what a clause says and not whether a case falls inside a defined term, which may explain why definitions barely change its score. In ContractNLI \citep{contractnli}, the largest accuracy change is 1.1 points. These public tasks use different questions and metrics from our diagnostic tests, so the small changes leave open how often benchmarks or real requests depend on definition boundaries. We designed the diagnostic sets to test those boundaries and not to estimate how common they are.
\section{A small model ranks difficult diagnostic questions}
\label{sec:predict}
\paragraph{Scoring questions with the proxy.}
To build an evaluation, we need to find the questions where a reader is at risk. We use a small open model, the proxy, to rank diagnostic questions. Let $z$ denote a question paired with the source occurrence of its term, and let $a^*$ be its keyed answer. Let $p_S^0(a^*\mid z)$ be the probability that the proxy assigns to the keyed option without the definition. The score is
\begin{equation}
R_S(z)=1-p_S^0(a^*\mid z).
\label{eq:risk}
\end{equation}
A high $R_S$ means that the proxy gives little probability to the keyed answer. If the proxy already favors the keyed answer, the question is probably easy. If it favors another option, larger readers may be misled as well. Scoring needs the answer key from the source but no response from the target reader. We then test whether this score ranks the errors that other readers make on the same questions. Because it needs the key, $R_S$ is an offline difficulty measure and cannot detect errors online when the correct answer is unknown. The confidence scores in Section~\ref{sec:detect} have no access to the key.

We also report $\Delta p_S(z)=p_S^d(a^*\mid z)-p_S^0(a^*\mid z)$ for a definition text $d$. This is the change in the proxy's probability for the keyed answer after we add that text. A different wording can give a different score.

\begin{table}[t]\centering\footnotesize\setlength{\tabcolsep}{4pt}
\caption{Reader error without and with the definition text, and AUROC for ranking errors without the definition. $R$ is the proxy score. Freq. is the better of term frequency and term rarity as a ranking score. The definition text is a drafted gloss, written from the source, for synthetic, Reddit, and DeFi, and extracted contract wording for CUAD. Rows differ in reader and question construction (Appendix~\ref{app:setting-design}), and CUAD is risk-stratified.}
\label{tab:prediction}
\begin{tabular}{@{}llrcccc@{}}\toprule
Corpus & Reader & $n$ & Error, no def. & Error, def. & $R$ AUROC [95\% CI] & Freq.\\\midrule
Synthetic & DeepSeek V4 Flash & 456 & 0.285 & 0.029 & 0.848 [0.811, 0.883] & 0.540\\
Synthetic & GPT-6-astra & 456 & 0.182 & 0.018 & 0.809 [0.763, 0.856] & 0.629\\
CUAD & GPT-5.6-terra & 1,500 & 0.213 & 0.160 & 0.815 [0.789, 0.840] & 0.507\\
Reddit & DeepSeek V4.1 Flash & 1,391 & 0.301 & 0.185 & 0.808 [0.785, 0.835] & 0.544\\
Reddit & GPT-6-astra & 1,391 & 0.146 & 0.119 & 0.720 [0.676, 0.761] & 0.588\\
DeFi & DeepSeek V4.1 Flash & 379 & 0.282 & 0.193 & 0.809 [0.763, 0.856] & 0.510\\
DeFi & GPT-5.6-terra & 379 & 0.230 & 0.148 & 0.792 [0.742, 0.839] & 0.589\\
DeFi & GPT-6-astra & 379 & 0.169 & 0.135 & 0.764 [0.704, 0.815] & 0.603\\
\bottomrule\end{tabular}\end{table}
\paragraph{Cross-reader error ranking.}
With answer keys, Qwen2.5-7B can rank the errors that other readers make on the same questions. On 456 shared synthetic questions, $R$ reaches 0.848 for DeepSeek V4 Flash and 0.809 for GPT-6-astra. It reaches 0.815 for GPT-5.6-terra on 1,500 stratified CUAD questions. The five exploratory Reddit and DeFi settings range from 0.720 to 0.809. Using the observed errors of another frontier reader gives a reference range of 0.640 to 0.892. Term frequency and rarity reach at most 0.63. On the synthetic questions, AUROC rises from 0.699 with the 0.5B proxy to 0.848 with the 7B proxy and 0.867 with the 14B proxy.

\paragraph{General difficulty.}
The score also ranks errors on terms that keep their usual meaning. In the synthetic corpus, it reaches 0.921 for DeepSeek V4 Flash and 0.846 for GPT-6-astra, based on 21 and 6 errors. On Reddit control words, the AUROCs are 0.842 and 0.815. Thus the score reflects general difficulty. We use it to choose which questions to review first before calling the target reader, and not to detect misreadings.

\begin{figure}[t]
\centering\includegraphics[width=.82\linewidth]{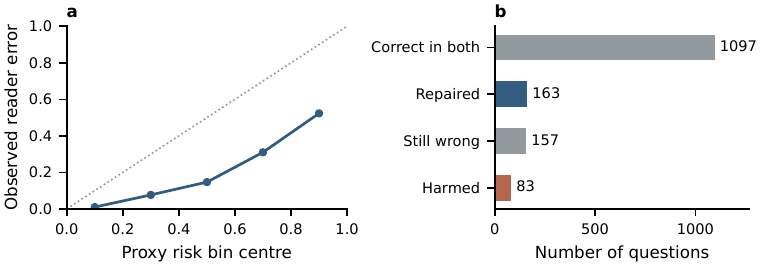}
\caption{CUAD validation with GPT-5.6-terra as the reader. We sample 300 questions from each score bin. (a) Observed error rises with proxy score. The dotted line shows where the two values would be equal. The score ranks errors and does not estimate their probability. (b) Outcomes after adding the extracted contract definition. More errors are fixed than correct answers are made wrong, and about half of the initial errors remain.}
\label{fig:strata}
\end{figure}

\paragraph{A scored contract frame.}
We score 6,529 questions from 303 CUAD contracts. We divide the scores into five bins and have GPT-5.6-terra answer 300 questions from each. The observed error rate rises from 1.0\% in the lowest bin to 52.3\% in the highest (Figure~\ref{fig:strata}). Reweighting to all scored questions gives a 13.2\% error rate and an AUROC of 0.846. The top two bins hold 23\% of the scored questions and 66\% of the estimated errors, so reviewers can prioritize questions within the collection.

\paragraph{Source-review sensitivity.}
An author used the available source to answer 27 questions without seeing the stored keys. The sample included 17 model-screen disagreements and ten stratified controls. On later review, we found support for 10 stored keys. We classified 14 questions as invalid or insufficiently supported by the source, and the other 3 as borderline. The sample gives extra weight to cases where the screening models disagree, so it cannot estimate the defect rate of the full dataset.

We removed the 14 problem items to see how much they affected the rankings. The largest change in $R$ AUROC across eight settings is 0.00203. The weighted CUAD estimate moves from 0.8456 to 0.8467, and the count of stable synthetic errors stays at 93 of 229. Most keys remain unreviewed.
\section{Implications}
\label{sec:discussion}
\paragraph{Separate rule use from evidence seeking.}
A model might know how to apply a definition without knowing when to look for one. It might also notice a gap and still fail to find the right rule. We directly test rule use. For gap detection, we have only indirect evidence from confidence and repeated answers, and we do not test a search agent with access to the source. If a passage gives no hint of a local exception, the reader may have no way to infer the correct scope from that passage alone. The practical question is when to look up a term, which depends on the available clues and the cost of unnecessary searches. Since the model's own signals give little warning, help has to come from outside.

\paragraph{Keep definitions tied to their sources.}
In the synthetic test, definitions based on the source help more than glosses based on the reader's prior knowledge. When storing a definition in a semantic layer, keep its source and scope, and keep the links to the clauses or exhibits that it refers to. This advice applies to software terminology \citep{se-terminology}, slang \citep{informal-language}, and organizational text. Automatic paraphrases need checking because they can fix some answers and make others worse. In our tests, confidence, sample agreement, and the contrastive-decoding setup we evaluated gave limited protection.

\paragraph{Evaluate against the full source.}
An answer can agree with a passage and still conflict with a definition elsewhere in the document. An evaluation should check both whether the answer fits the passage and whether it is correct under the full source. Our diagnostic questions target this difference, and a small proxy with answer keys can help choose which of them to review first. The public-task results do not tell us how often it matters in ordinary use.
\section{Limitations and conclusion}
\label{sec:conclusion}
Models drafted most of the diagnostic questions and their keys. The questions show that errors occur, but they are multiple choice and were not written by real users. The source review found invalid questions and questions whose keys the available source did not support, and removing those items barely changes the rankings. The remaining keys have not all been validated. We treat Reddit and DeFi as exploratory. The proxy score needs the answer key, so it cannot flag risky live requests. Our results also leave open how models behave when they can ask for clarification, abstain, or search the source, and how often the problem arises in real user traffic.

We found three things. First, a missing local definition can lead a reader to repeat the same wrong answer, and the confidence signals we test give little warning of these errors. Second, the source definition substantially improves accuracy on the controlled synthetic task, and a prior gloss does not. In contracts, definitions fix some errors but introduce others, and some decisions need additional evidence. Third, a small proxy with answer keys can help reviewers find difficult evaluation questions. These results call for tests tied to source definitions and for retrieval that respects their scope.
\label{main:end}
\bibliography{references}
\bibliographystyle{iclr2027_conference}
\clearpage\appendix
\section{Data and question construction}
\label{app:sources}
We draft diagnostic questions from source passages and local definitions. Each question has a set of options and a key based on the source. The keys let us measure reader behavior. They do not represent expert decisions on every case. The supplementary inventory records which task version we use. It also distinguishes results for individual questions from means for a term.

\paragraph{Synthetic vocabulary.}
The synthetic corpus uses invented vocabulary for a trading desk and a restaurant group. We set each term's class and local definition before generating text. The diagnostic intervention adds a gloss drafted from the source.

The archived uncertainty experiment has 763 questions from 197 nonempty term records. We exclude two questions where the source and definition conflict, leaving 761 for analysis. The archived overlap with the 7B proxy has 457 questions from 119 records. One of the affected questions is in that overlap, so 456 remain. The usage-based inline test contains 965 questions. We vary frequency by design. The rarity comparisons thus describe the synthetic corpus rather than a natural distribution.

\paragraph{CUAD.}
CUAD contains 510 contracts. We extract 2,174 term records from 303 contracts and identify 24,880 candidate occurrences. After drafting questions and applying the per-term limits, we score 6,529 questions. The stratified validation sample has 1,500 questions from 1,012 term records in 215 contracts. The separate definition-anchored task has 671 questions on 187 terms. We analyze the two sets separately because we built them in different ways.

The introductory example is record \texttt{cuadfull:379:Trademarks}, question at context index 2. Its definition restricts Trademarks to the trademarks and trade names identified in Exhibit~A. The usage requires the distributor to notify the manufacturer of unauthorized use of the Trademarks, so the notification duty covers only the listed names.

\paragraph{Reddit.}
The Reddit data have 393 term records. Of these, 352 have questions. The 1,391 retained questions cover 35 communities. The retrieval workload includes 851 questions about glossary terms and 540 control questions. All 393 stored \texttt{gloss\_used} fields contain drafted glosses. For the 218 glossary terms that have a source entry, none of those glosses matches \texttt{gloss\_gold} exactly. A control term need not have a source entry. A test using \texttt{gloss\_used} measures the effect of the rewritten text.

\paragraph{DeFi.}
The DeFi set tests technical vocabulary on 379 questions from 97 nonempty term records. All 98 candidate glosses were written by a model. They differ from the recorded source glosses. The official model cards list February 16, 2026 as the knowledge cutoff for GPT-5.6-terra and April 30, 2026 for GPT-6-astra \citep{openai-gpt56-terra-model,openai-gpt6-astra-model}.

Some files in the documentation snapshot have later dates. Each term links a median of 18 documents, and we cannot verify when the rule needed for a given answer first appeared. Uniswap v4 and Unichain were public in 2025 \citep{uniswap-v4-launch,unichain-launch}. The GPT readers' $R$ AUROCs of 0.792 and 0.764 describe error ranking on these questions. The analysis does not assume that the models had no access to earlier versions during training. It cannot rule out memorization.

\paragraph{Question and answer-key validity.}\label{app:source-screen}
We use a fixed SHA-256 rule to select 20 archived questions per corpus before checking any new answers against the stored keys. For Reddit, we draw half from the ordinary question set and half from a fixed set on which strategies disagree. The 10/10 split does not represent the natural frequency of either group.

DeepSeek V4.1 Flash and Claude Sonnet 5 each see the same source cards and excerpts chosen from the linked documents using the question alone. Both run without explicit reasoning. DeepSeek uses temperature zero and a 550-token limit. Sonnet 5 rejects temperature zero, so we use the provider default with a 1,000-token limit. All 80 Sonnet responses finish normally. The host does not provide response IDs. Neither model sees the stored key or the reader outcomes. The long contracts and DeFi documents appear as excerpts. For Reddit, the archive retains glossary entries and selected comments rather than every original post in full.

\begin{table}[h]\centering\small
\caption{Model source screening on 20 questions per corpus. ``Both key'' means both models gave a unique source-supported answer matching the stored key. ``Both other'' means they agreed on another answer. ``Split answer'' means two supported but different answers. ``One supported'' means one model judged the source sufficient and the other did not. ``Neither supported'' includes insufficient or ambiguous evidence. These are screening categories, not expert answer-key error rates.}
\label{tab:source-screen}
\begin{tabular}{@{}lrrrrr@{}}\toprule
Corpus & Both key & Both other & Split answer & One supported & Neither supported\\\midrule
Synthetic & 18 & 0 & 0 & 1 & 1\\
CUAD & 13 & 1 & 1 & 4 & 1\\
Reddit & 3 & 0 & 0 & 3 & 14\\
DeFi & 5 & 0 & 0 & 7 & 8\\\bottomrule
\end{tabular}
\end{table}

\paragraph{Human source review.}
An author answers 27 of the 80 selected questions from the available source before seeing the stored keys. Seventeen come from model-screen disagreements. The other ten are sampled by corpus and screen category. The author chooses an option on 11 questions. The other 16 are undetermined. A different answer does not always mean the key is wrong. In some cases, the local definition supports the stored key. Table~\ref{tab:review} gives the results after checking disagreements against the source with the key visible. This later check is not a second blinded annotation, so we do not report an agreement coefficient.
\begin{table}[h]\centering\footnotesize
\caption{Source review of 27 selected diagnostic questions: 17 model-screen disagreements and ten stratified controls. Counts describe this enriched sample, not corpus-wide defect rates.}
\label{tab:review}
\begin{tabular}{@{}lcccc@{}}\toprule
Corpus & Key supported & Invalid or source-insufficient & Borderline & Reviewed\\\midrule
Synthetic & 3 & 1 & 0 & 4\\
CUAD & 2 & 6 & 1 & 9\\
Reddit & 3 & 1 & 1 & 5\\
DeFi & 2 & 6 & 1 & 9\\\midrule
Total & 10 & 14 & 3 & 27\\\bottomrule
\end{tabular}\end{table}

The review found several kinds of defect. One question needs a redacted exhibit to establish a manufacturer's duty. Another gives only A/B/C/D placeholders as options. A removal question treats removal as mandatory, although the contract merely permits it. A blocked swap can also have two different causes. In one CUAD item, the stored option lists user guides and technical manuals as rights that were not granted. The contract explicitly grants rights to Documentation. We remove that item rather than apply scores for the old key to a new answer.

A disagreement can also be resolved without changing the key. For example, the community entries for Fill, Burst, and Standard can resolve the constructed Reddit scenarios even if no archived post describes the exact case. We check the source before deciding whether a disagreement is a key error.
\begin{table}[h]\centering\footnotesize
\caption{Deletion sensitivity after removing the 14 reviewed questions classified as clearly invalid or source-insufficient. Other questions retain their historical keys. This is a local sensitivity analysis, not a validated-only re-estimate. $\Delta p$ AUROC targets repair among observed bare errors on the same question.}
\label{tab:validity-sensitivity}
\begin{tabular}{@{}p{4.0cm}rcc@{}}\toprule
Setting & Items retained & $R$ error AUROC & Conditional repair $\Delta p$ AUROC\\\midrule
Synthetic / DeepSeek V4 Flash & 455/456 & .848 $\to$ .847 & .983 $\to$ .983\\
Synthetic / GPT-6-astra & 455/456 & .809 $\to$ .809 & .962 $\to$ .962\\
CUAD / GPT-5.6-terra & 1,494/1,500 & .815 $\to$ .817 & .712 $\to$ .714\\
Reddit / DeepSeek V4.1 Flash & 1,390/1,391 & .808 $\to$ .808 & .858 $\to$ .860\\
Reddit / GPT-6-astra & 1,390/1,391 & .720 $\to$ .720 & .812 $\to$ .812\\
DeFi / DeepSeek V4.1 Flash & 373/379 & .809 $\to$ .808 & .863 $\to$ .863\\
DeFi / GPT-5.6-terra & 373/379 & .792 $\to$ .791 & .794 $\to$ .794\\
DeFi / GPT-6-astra & 373/379 & .764 $\to$ .764 & .846 $\to$ .852\\\bottomrule
\end{tabular}
\end{table}

We remove the 14 reviewed items with clear defects or insufficient sources. Across the eight matched settings, the largest change in $R$ error-ranking AUROC is 0.00203. The largest change in same-question repair $\Delta p$ AUROC is 0.00601. The weighted CUAD $R$ AUROC moves from 0.8456 to 0.8467. The synthetic count stays at 93 repeated wrong answers among 229 greedy errors. Removing the three borderline items as well has little effect on the estimates.

The 17 targeted items and ten stratified controls do not form a random sample of all archived questions. Their category counts cannot estimate defect rates. The deletion check does not validate the remaining unreviewed keys. Of the original 80 questions, 24 remain unresolved because neither model found a unique answer supported by the source. We do not count them as wrong keys.

\paragraph{Definition version by CUAD task.}
For the 1,500-question CUAD set, the proxy and reader receive \texttt{def\_raw}, the extracted contract wording. The separate 671-question task has a drafted gloss, \texttt{glossC}, and a source definition, \texttt{glossG}. All 204 term records in the smaller package mark \texttt{gloss\_used} as drafted and different from the source entry. In the archived prompts, all 671 \texttt{glossC} notes match the drafted text. All 671 \texttt{glossG} notes match the original contract text. The two CUAD tasks thus test different definition interventions.

\paragraph{Eight nonexchangeable reader settings.}\label{app:setting-design}
Table~\ref{tab:setting-design} lists the reader, the definition it received, and what the question writer saw. The synthetic writer saw usage examples without the source definition. In Reddit and DeFi, the source entry helped the writer draft the question and key. The reader later received a different, drafted gloss. For CUAD, the writer saw an extracted contract definition and one use of the term. Within each corpus, the readers answered the same questions. The corpora differ in how we drafted questions and which definition we supplied. The rows are thus not repeated tests of a single setup.

\begin{table}[h]\centering\footnotesize
\caption{Reader settings in Table~\ref{tab:prediction}. Synthetic DeepSeek calls used V4 Flash. Later calls used V4.1 Flash. DeepSeek documents the API IDs \texttt{deepseek-v4-flash} and \texttt{deepseek-flash}, respectively. ``Source entry'' describes drafting input and does not imply that the reader's \texttt{glossC} arm used that entry.}
\label{tab:setting-design}
\begin{tabular}{@{}>{\raggedright\arraybackslash}p{1.5cm}>{\raggedright\arraybackslash}p{3.0cm}>{\raggedright\arraybackslash}p{3.4cm}>{\raggedright\arraybackslash}p{4.7cm}@{}}\toprule
Corpus & Reader & Definition given to reader & Question drafting input\\\midrule
Synthetic & DeepSeek V4 Flash & Drafted gloss (\texttt{glossC}) & Invented-term usage contexts. No source definition in drafting prompt\\
Synthetic & GPT-6-astra & Drafted gloss (\texttt{glossC}) & Same synthetic questions\\
CUAD & GPT-5.6-terra & Source definition (\texttt{def\_raw}) & One definition and one contract usage per question\\
Reddit & DeepSeek V4.1 Flash & Drafted gloss (\texttt{glossC}) & Community glossary entry and comment contexts\\
Reddit & GPT-6-astra & Drafted gloss (\texttt{glossC}) & Same Reddit questions\\
DeFi & DeepSeek V4.1 Flash & Drafted gloss (\texttt{glossC}) & Source glossary entry and documentation contexts\\
DeFi & GPT-5.6-terra & Drafted gloss (\texttt{glossC}) & Same DeFi questions\\
DeFi & GPT-6-astra & Drafted gloss (\texttt{glossC}) & Same DeFi questions\\\bottomrule
\end{tabular}
\end{table}

\paragraph{Designed-class control for mechanism specificity.}\label{app:synthetic-control}
The synthetic vocabulary assigns each term to one of three classes before any diagnostic question is generated. Redefined common words have a usual meaning that conflicts with the local one. Invented words have no usual meaning. Usual-meaning words have a usual meaning that matches the local one. Table~\ref{tab:synthetic-control} evaluates $R$ against reader bare errors separately in each class, with term-record bootstrap intervals. The score ranks errors for invented words and for usual-meaning words, where no conflict was designed into the term. The error-ranking performance is thus not specific to meaning conflict. The classes are design labels and not independent diagnoses of each error, and the GPT-6-astra row for usual-meaning words has only six errors.

\begin{table}[h]\centering\small
\caption{Error-ranking $R$ by designed synthetic term class. Reader cells give errors, AUROC [95\% CI]. Both readers answer the same 456 questions. Intervals resample term records.}
\label{tab:synthetic-control}
\begin{tabular}{@{}lrrr@{}}\toprule
Class & Questions & DeepSeek V4 Flash & GPT-6-astra\\\midrule
Redefined common words & 156 & 69, 0.803 [0.739, 0.864] & 57, 0.777 [0.697, 0.854]\\
Invented words & 158 & 40, 0.779 [0.704, 0.849] & 20, 0.720 [0.629, 0.819]\\
Usual-meaning words & 142 & 21, 0.921 [0.870, 0.966] & 6, 0.846 [0.710, 0.993]\\\bottomrule
\end{tabular}
\end{table}

\paragraph{Reddit non-glossary controls.}
The Reddit workload also contains 540 questions attached to terms outside the community glossary. DeepSeek V4.1 Flash has 0.235 bare-error rate and $R$ AUROC 0.842 on these controls, versus 0.343 and 0.781 on 851 glossary-term questions. GPT-6-astra has corresponding control values 0.143 and 0.815, versus 0.148 and 0.659 on glossary terms. The non-glossary terms are a useful difficulty reference, but absence from the collected glossary does not prove absence of every local sense. The separate source audit also finds substantial Reddit question-answerability uncertainty.

\paragraph{Other readers as a difficulty reference.}\label{app:reader-difficulty}
One way to estimate shared difficulty is to see which questions another reader gets wrong without the definition. Table~\ref{tab:reader-difficulty} compares $R$ with a binary score from one reader or an error-count score from two. This comparison requires answers from another frontier model. It is neither a cheap deployment baseline nor an upper bound. The scores differ in resolution. Both can respond to the same local-meaning error. We did not rescore the questions with the source occurrence or target term removed, so the comparison cannot isolate a semantic component of $R$.

\begin{table}[h]\centering\footnotesize
\caption{AUROC for predicting a target reader's bare errors on the same questions. ``Other readers'' is one observed reader error or the mean of two other reader errors. All models share the historical question and answer key.}
\label{tab:reader-difficulty}
\begin{tabular}{@{}llrcc@{}}\toprule
Corpus & Target reader & $n$ & Proxy $R$ & Other readers\\\midrule
Synthetic & DeepSeek V4 Flash & 456 & .848 & .722\\
Synthetic & GPT-6-astra & 456 & .809 & .804\\
Reddit & DeepSeek V4.1 Flash & 1,391 & .808 & .640\\
Reddit & GPT-6-astra & 1,391 & .720 & .736\\
DeFi & DeepSeek V4.1 Flash & 379 & .809 & .750\\
DeFi & GPT-5.6-terra & 379 & .792 & .867\\
DeFi & GPT-6-astra & 379 & .764 & .892\\\bottomrule
\end{tabular}
\end{table}

\paragraph{Stability and key-free signals by term class.}\label{app:signals-by-class}
Table~\ref{tab:signals-by-class} splits the synthetic stability and detection results by the designed term class. Stability columns use the 761-question sampling frame for DeepSeek V4 Flash. Detection columns use the 456 questions shared with the proxy, where error counts are smaller. Sample disagreement and option entropy give identical rankings on these ten-sample votes, so only disagreement is shown. The key-free proxy score is one minus the proxy's largest option probability. The released scoring prompt reproduces the stored keyed probabilities to within 0.005. Intervals resample term records.

\begin{table}[h]\centering\scriptsize\setlength{\tabcolsep}{3pt}
\caption{Synthetic stability and error-detection AUROC by term class for DeepSeek V4 Flash. ``Same wrong'' counts greedy errors repeated identically in all ten samples. ``Majority wrong'' counts errors that survive majority voting.}
\label{tab:signals-by-class}
\resizebox{\linewidth}{!}{\begin{tabular}{@{}lrrrrcccc@{}}\toprule
 & \multicolumn{3}{c}{Stability (761 questions)} & \multicolumn{5}{c}{Detection AUROC (456 questions)}\\\cmidrule(lr){2-4}\cmidrule(lr){5-9}
Class & Errors & Same wrong & Majority wrong & Errors & Disagreement & Stated conf. & Proxy, no key & Proxy $R$\\\midrule
Redefined common words & 168/461 & 74 (44\%) & 110 (65\%) & 69 & 0.557 [0.492, 0.622] & 0.463 [0.376, 0.554] & 0.537 [0.439, 0.636] & 0.803 [0.738, 0.864]\\
Invented words & 40/158 & 12 (30\%) & 23 (58\%) & 40 & 0.641 [0.570, 0.720] & 0.715 [0.637, 0.785] & 0.718 [0.615, 0.816] & 0.779 [0.704, 0.848]\\
Usual-meaning words & 21/142 & 7 (33\%) & 9 (43\%) & 21 & 0.603 [0.517, 0.692] & 0.644 [0.552, 0.730] & 0.810 [0.710, 0.897] & 0.921 [0.871, 0.965]\\
All & 229/761 & 93 (41\%) & 142 (62\%) & 130 & 0.616 [0.579, 0.656] & 0.617 [0.558, 0.681] & 0.689 [0.632, 0.750] & 0.848 [0.812, 0.881]\\
\bottomrule\end{tabular}}\end{table}

\begin{table}[h]\centering\footnotesize
\caption{Bare error and proxy $R$ AUROC by term class. Synthetic classes are fixed by the corpus design. Reddit glossary terms have a community entry, and control words are ordinary words from the same communities without one.}
\label{tab:classes}
\begin{tabular}{@{}llrcccc@{}}\toprule
 & & & \multicolumn{2}{c}{DeepSeek Flash (V4/4.1)} & \multicolumn{2}{c}{GPT-6-astra}\\\cmidrule(lr){4-5}\cmidrule(lr){6-7}
Corpus & Term class & $n$ & Error & $R$ AUROC & Error & $R$ AUROC\\\midrule
Synthetic & Redefined common words & 156 & 0.442 & 0.803 & 0.365 & 0.777\\
Synthetic & Invented words & 158 & 0.253 & 0.779 & 0.127 & 0.720\\
Synthetic & Usual-meaning words & 142 & 0.148 & 0.921 & 0.042 & 0.846\\
Reddit & Glossary term & 851 & 0.343 & 0.781 & 0.148 & 0.659\\
Reddit & Control word & 540 & 0.235 & 0.842 & 0.143 & 0.815\\\bottomrule
\end{tabular}\end{table}

\section{Scoring and statistical methods}
\label{app:statistics}
\paragraph{Answer-label scores.}
The diagnostic proxy normalizes scores across answer letters. The seven proxy tokenizers represent both the plain and space-prefixed A/B/C/D labels with one token per spelling, without cross-letter token collisions. The historical estimator takes the higher score over these two spellings for each letter. This convention defines the proxy score and is not a claim of calibrated reader confidence. Model and tokenizer revisions appear in Appendix~\ref{app:models}.

The signed probability change is $\Delta p_S(z)=p_S^d(a^*\mid z)-p_S^0(a^*\mid z)$ for a definition text $d$ (Section~\ref{sec:predict}). It is indexed by the exact definition text, its scope and its version. A mean log-score contrast, stored as \texttt{delta\_mc} in the supplementary data, is a separate statistic and is not substituted for $\Delta p$.

\paragraph{Pairing and evaluation units.}
Prediction comparisons match record IDs and question stems, with context indices for the CUAD library. Reader records specify answer keys and option order. The main table reports question-level outcomes. Term-level Sonnet summaries have a different unit and remain in the supplementary inventory.

\paragraph{Definition-anchored CUAD task.}
On 671 matched questions, accuracy is 0.772 without a definition. It reaches 0.903 with the drafted \texttt{glossC} and 0.982 with the original contract text in \texttt{glossG}. The latter two conditions differ in correctness on 57 items. For error ranking, AUROC is 0.858 for $R$ and 0.666 for $\Delta p$ computed with the drafted gloss. These proxy scores concern the drafted text. The 1,500-question usage validation is a separate test.

\paragraph{Stratified validation.}
The five CUAD risk strata contain 2,529, 1,430, 1,085, 980, and 505 scored questions. We select 300 questions from each stratum using a seeded shuffle (seed 42). An item's weight is its stratum size divided by 300. We apply the weights to pairs of correct and incorrect items to compute error AUROC. Across the 6,529 scored questions, the weighted failure rate is 13.2\%. The weighted AUROC is 0.846 for $R$ and 0.500 for $\Delta p$. These weights do not cover occurrences that were excluded during question construction.

\paragraph{Uncertainty intervals.}
We use 2,000 cluster-bootstrap resamples for the main intervals. We use the same resample for both arms in paired comparisons. A cluster is a contract in CUAD or a community in Reddit. For synthetic and DeFi questions, we resample term records. The intervals describe variation across the observed clusters. They do not cover changes in organization, question writer, or repeated model calls. We report subgroup and threshold results as exploratory.

The Synthetic / GPT-6-astra and DeFi / DeepSeek V4.1 Flash rows in Table~\ref{tab:prediction} round to the same $R$ point estimate and interval at three decimal places. Their unrounded values differ: 0.8094577 with interval $[0.7627638,0.8560215]$ and 0.8093046 with $[0.7625671,0.8555733]$, respectively. The printed values coincide only after rounding. The manuscript uses the same bootstrap procedure for both rows.

\paragraph{Repair prediction.}
Table~\ref{tab:repair} evaluates prediction of a wrong-to-correct transition after a definition. Its final column reports a paired interval for the difference between the two scores on the separate bare-error target.
\begin{table}[t]\centering\footnotesize\setlength{\tabcolsep}{4pt}
\caption{Prediction of observed repair. The last column concerns the separate bare-error target and gives a paired ranking difference. $R$ and $\Delta p$ are algebraically coupled (Appendix~\ref{app:coupling}).}
\label{tab:repair}
\begin{tabular}{@{}p{4.5cm}rccc@{}}\toprule
Setting & $n$ & \shortstack{Repair AUROC\\$R$} & \shortstack{Repair AUROC\\$\Delta p$} & \shortstack{Error AUROC difference\\$\Delta p-R$, 95\% CI}\\\midrule
Synthetic / DeepSeek V4 Flash & 456 & 0.833 & 0.810 & [-0.168, -0.046]\\
Synthetic / GPT-6-astra & 456 & 0.797 & 0.716 & [-0.259, -0.066]\\
CUAD library / GPT-5.6-terra & 1500 & 0.760 & 0.595 & [-0.395, -0.302]\\
Reddit / DeepSeek V4.1 Flash & 1391 & 0.744 & 0.709 & [-0.323, -0.244]\\
Reddit / GPT-6-astra & 1391 & 0.725 & 0.621 & [-0.364, -0.213]\\
DeFi / DeepSeek V4.1 Flash & 379 & 0.769 & 0.686 & [-0.445, -0.251]\\
DeFi / GPT-5.6-terra & 379 & 0.792 & 0.556 & [-0.502, -0.327]\\
DeFi / GPT-6-astra & 379 & 0.754 & 0.590 & [-0.510, -0.324]\\
\bottomrule\end{tabular}\end{table}
\paragraph{Algebraic coupling of error risk and definition response.}\label{app:coupling}
Let $p_0$ and $p_d$ be the proxy probabilities assigned to the keyed option without and with the supplied definition. By construction, $R=1-p_0$ and $\Delta p=p_d-p_0=R+p_d-1$. Because $0\leq p_d\leq1$, the two scores satisfy $R-1\leq\Delta p\leq R$. The scores share the same $p_0$ term. Their ranking comparison should not be treated as evidence from independent signals. The paired bootstrap interval in Table~\ref{tab:repair} compares their rankings on the same bare-error target. Pairing handles shared sampling variation but does not remove this structural dependence. It is not evidence that $\Delta p$ isolates a distinct cause of error.

\begin{table}[h]\centering\footnotesize
\caption{Why $\Delta p$ can rank bare-reader errors below chance. Means are listed as bare-error / bare-correct items. The final column compares only error-correct pairs inside the same proxy-$R$ decile, a descriptive approximate control for the common $p_0$ term. It is not a causal decomposition.}
\label{tab:dp-coupling}
\begin{tabular}{@{}lcccc@{}}\toprule
Setting & $\Delta p$ error AUROC & Mean $p_d$ & Mean $\Delta p$ & Within-$R$-decile AUROC\\\midrule
Synthetic / DeepSeek V4 Flash & 0.745 & .804 / .924 & .496 / .246 & .485\\
Synthetic / GPT-6-astra & 0.656 & .751 / .920 & .450 / .287 & .431\\
CUAD / GPT-5.6-terra & 0.464 & .407 / .714 & .139 / .141 & .398\\
Reddit / DeepSeek V4.1 Flash & 0.520 & .440 / .675 & .163 / .131 & .409\\
Reddit / GPT-6-astra & 0.429 & .405 / .639 & .105 / .147 & .359\\
DeFi / DeepSeek V4.1 Flash & 0.458 & .410 / .670 & .116 / .130 & .411\\
DeFi / GPT-5.6-terra & 0.380 & .356 / .668 & .064 / .144 & .307\\
DeFi / GPT-6-astra & 0.342 & .330 / .651 & .034 / .144 & .273\\\bottomrule
\end{tabular}
\end{table}

For the bare-error target, a $\Delta p$ AUROC below 0.5 means that observed errors tend to have a smaller change in probability after the definition. Consider DeFi / GPT-6-astra. Mean $R$ is 0.704 for bare errors and 0.494 for bare-correct answers. Mean post-definition support $p_d$ is 0.330 for bare errors and 0.651 for bare-correct answers. This latter difference more than offsets the difference in $R$. Mean $\Delta p$ is thus 0.034 for bare errors and 0.144 for bare-correct answers. The CUAD means almost cancel, but the pairwise ranking still gives an AUROC of 0.464.

Ranking bare errors is different from ranking repairs among bare errors. A score can perform weakly on the first task and still help on the second. The within-decile results are descriptive. They do not control question difficulty beyond $R$, how the questions were drafted, or the quality of the keys.

The more specific target restricts evaluation to a reader's bare failures and asks which of those errors the same definition fixes. Table~\ref{tab:conditional-repair} gives direct and within-term sibling estimates for the recorded intervention text. Conditional results cannot by themselves guide pre-answer selection, because bare failure is then unknown.

\paragraph{Question-independent meaning difference.}\label{app:meaning-difference}
We measure how far the source definition is from the model's prior gloss. Without domain context, the distance is $1-\cos(e(d_{\mathrm{source}}),e(g_{\mathrm{prior}}))$. Here $e$ is the reported BAAI/bge-base-en-v1.5 sentence encoder \citep{bge}. We compute a second distance using the prior gloss elicited with a domain prompt. We also ask the model whether the term has a special use in the corpus. Its relation labels include none, narrower, and shifted. None of these measures uses a diagnostic question.

For comparison, we report three scores that do use diagnostic questions. The first compares the drafted gloss with the elicited prior gloss, averaged by term. The other two are term means of $\Delta p$ and $R$. We cannot compute a distance when the model gives no prior gloss.

\begin{table}[h]\centering\small
\caption{Distinguishing designed redefinitions from usual-meaning terms in the synthetic vocabulary, after excluding one source-definition-conflicted term. Each score uses all available terms for that score. Coverage differs and point estimates should not be ranked as if measured on one common sample. The class label was fixed by construction.}
\label{tab:term-difference-synthetic}
\begin{tabular}{@{}lccc@{}}\toprule
Term score & Uses questions? & Terms (shifted / consistent) & Class AUROC\\\midrule
No-context definition distance & No & 66 (38 / 28) & 0.946\\
Domain-prior definition distance & No & 157 (118 / 39) & 0.977\\
Model self-report & No & 79 (40 / 39) & 0.984\\
Drafted-local/prior answer contrast & Yes & 78 (40 / 38) & 0.917\\
Mean $\Delta p$ & Yes & 78 (40 / 38) & 0.783\\
Mean $R$ & Yes & 78 (40 / 38) & 0.817\\\bottomrule
\end{tabular}
\end{table}

The strong synthetic separation is a positive control for a deliberately designed semantic contrast, not an estimate of real-corpus misreading. To compare a source-definition distance with reader behavior, we use only the archived reader arm that supplied that same source definition, \texttt{glossG}. A term's repair rate is the fraction of its historical questions answered wrongly without a definition and correctly with \texttt{glossG}. This term-level outcome uses a median of four questions per term. The automatically drafted \texttt{glossC} arm would compare a different intervention text and is excluded from this alignment.

\begin{table}[h]\centering\small
\caption{Spearman association of question-independent term scores with source-definition repair rate. Parentheses give available terms. Distance confidence intervals resample term records. There is no aligned \texttt{glossG} reader arm for this 204-term CUAD package. Only 54 terms have an uncontextualized prior gloss.}
\label{tab:term-difference-repair}
\begin{tabular}{@{}lccc@{}}\toprule
Corpus & No-context distance [95\% interval] & Domain-prior distance & Self-report\\\midrule
Synthetic & $0.452$ $[0.260,0.602]$ (65) & $0.300$ (186) & $0.465$ (118)\\
Reddit & $0.181$ $[0.046,0.303]$ (211) & $0.095$ (217) & $0.166$ (217)\\
DeFi & $-0.087$ $[-0.361,0.186]$ (57) & $0.101$ (64) & $-0.237$ (64)\\\bottomrule
\end{tabular}
\end{table}

For Reddit, we use proxy scores from the same community entry that readers saw. Across 209 terms with 418 A questions, the distance from the prior gloss without domain context has a Spearman correlation of $0.188$ with the term's mean source-definition $\Delta p$. Resampling term records gives an interval of $[0.047,0.323]$. This weak positive association does not tell us which future question will fall on a definition boundary. Both analyses use archived questions drafted by models. The source review also leaves substantial uncertainty about answerability on Reddit and DeFi.

Model-assigned relation labels also give weak and inconsistent repair separation on the source-definition arm. Among Reddit glossary terms, 186 labeled narrower or shifted have mean repair rate 0.292, versus 0.266 for 31 labeled none. The term-level repair AUROC is 0.531. In DeFi, the corresponding rates are 0.227 versus 0.333 over 32 terms per group, with AUROC 0.411. These are model-assigned lexical relations, not independent annotations of the three situations in Table~\ref{tab:cases}. A single text distance cannot by itself tell whether a use involves meaning conflict, an unspecified local scope, or missing information elsewhere. The synthetic reader-aligned text pairs and embedding script are retained, and the other source/prior texts remain in the archived weights. The exact BGE revision and embedding vectors are not, so exact bitwise reproduction of the cosine values is not guaranteed. The reported associations and source-aligned reader outcomes are reproducible from the archived scores.

\paragraph{Sampling consistency.}
The archived synthetic uncertainty test has a greedy response and ten sampled responses per question. There are 7,630 sampled letters in total. After excluding the two questions with conflicting sources, we use 761 questions and 7,610 sampled letters. Of the 229 greedy errors, 154 yield ten identical sampled answers. In 93 cases the repeated answer is wrong. In 61 it is correct. The majority vote remains wrong on 142 of the 229 items. The corrected comparison shared with the proxy covers 456 questions. We compute entropy over option votes. We treat stated confidence as a response measurement rather than a calibrated probability.

\begin{table}[t]\centering\footnotesize
\caption{Predicting which of a reader's errors the definition text repairs, as AUROC for ranking repaired above unrepaired errors among errors made without the definition. $R$ and $\Delta p$ use the same diagnostic question in the middle columns. The last column uses other scored questions of the same term record, with the available failed-question count in parentheses. CUAD uses the full proxy-scored frame for sibling estimates. Other rows use their available scored siblings. The CUAD 95\% interval is $[0.494,0.661]$.}
\label{tab:conditional-repair}
\begin{tabular}{@{}p{4.4cm}rccc@{}}\toprule
Setting & Bare failures & $R$ & $\Delta p$ & Other-item $\Delta p$ ($n$)\\\midrule
Synthetic / DeepSeek V4 Flash & 130 & 0.445 & 0.983 & 0.914 (130)\\
Synthetic / GPT-6-astra & 83 & 0.421 & 0.962 & 0.976 (83)\\
CUAD / GPT-5.6-terra & 320 & 0.462 & 0.712 & 0.581 (190)\\
Reddit / DeepSeek V4.1 Flash & 419 & 0.453 & 0.858 & 0.824 (418)\\
Reddit / GPT-6-astra & 203 & 0.554 & 0.812 & 0.864 (203)\\
DeFi / DeepSeek V4.1 Flash & 107 & 0.501 & 0.863 & 0.850 (107)\\
DeFi / GPT-5.6-terra & 87 & 0.568 & 0.794 & 0.799 (87)\\
DeFi / GPT-6-astra & 64 & 0.476 & 0.846 & 0.874 (64)\\
\bottomrule\end{tabular}\end{table}

\section{Definition interventions}
\label{app:interventions}
\begin{table}[h]\centering\footnotesize
\caption{DeepSeek Flash readers on matched questions under four conditions: V4 for synthetic and V4.1 for Reddit and DeFi. Cell values are accuracy, followed by repairs/harms relative to bare. \texttt{glossG} is the source definition, \texttt{glossC} the drafted gloss, and \texttt{glossM} the prior gloss, elicited under a domain prompt without the source examples. The synthetic row excludes two source-conflicted questions. Reddit and DeFi drafters saw the source definition.}
\label{tab:definition-arms}
\begin{tabular}{@{}lrrrrr@{}}\toprule
Corpus & $n$ & Bare & Source \texttt{glossG} & Drafted \texttt{glossC} & Prior \texttt{glossM}\\\midrule
Synthetic & 761 & .699 & .966 (209/6) & .949 (196/6) & .708 (86/79)\\
Reddit glossary & 851 & .657 & .932 (243/9) & .790 (171/58) & .616 (82/117)\\
DeFi source-entry & 250 & .672 & .940 (70/3) & .760 (42/20) & .740 (40/23)\\\bottomrule
\end{tabular}
\end{table}

\paragraph{Lexical shortcuts in generated questions.}\label{app:lexical-cues}
A question can share words with a gloss without quoting it. To screen for this cue, we remove the target term and common English words. For each option, we count the fraction of content words that occur in the exact drafted \texttt{glossC}. The rule selects the option with the highest fraction only if there is no tie. For the synthetic corpus, the \texttt{raw\_outputs} glosses match all 197 recovered reader term prompts. A different input file with the same name does not match. Table~\ref{tab:lexical-cues} shows how often the rule picks the stored key. As a baseline, we shuffle glosses among terms within a corpus. The reader still sees the original question in the subset that the rule does not flag. Other cues can remain in those questions.

\begin{table}[h]\centering\footnotesize
\caption{Lexical-cue screen using the exact \texttt{glossC} text supplied to the archived DeepSeek Flash readers (V4 for synthetic, V4.1 for Reddit and DeFi). Synthetic values use the reader-aligned \texttt{raw\_outputs/v1\_weights.jsonl}, rather than a different same-named input copy. Error rates are bare $\to$ drafted-gloss.}
\label{tab:lexical-cues}
\begin{tabular}{@{}lrrrrcc@{}}\toprule
Corpus & $n$ & Cue & Shuffle & No-cue $n$ & No-cue errors & No-cue AUROC: $R$ / repair-$\Delta p$\\\midrule
Synthetic & 456 & .686 & .067 & 143 & .308 $\to$ .091 & .879 / .950\\
Reddit & 1,391 & .185 & .032 & 1,133 & .296 $\to$ .210 & .810 / .844\\
DeFi & 379 & .274 & .086 & 275 & .284 $\to$ .236 & .803 / .791\\\bottomrule
\end{tabular}
\end{table}

The high synthetic cue rate means the full-sample 0.285 $\to$ 0.029 error reduction is partly exposed to an option-matching shortcut. The no-cue subset still shows a substantial change, but its definition arm may contain unmeasured semantic or structural cues. Reddit and DeFi have lower lexical-cue rates, yet their question writers saw source definitions and their answer-key audits have additional limitations.

\paragraph{Definition arms without lexical cues.}
For the four-arm synthetic comparison, the same warning rule is applied separately to the source definition, the drafted gloss, and the model's own gloss. The rule flags 45.2\%, 63.1\%, and 29.3\% of the 761 questions for the three texts. On the 180 questions flagged by none of them, DeepSeek V4 Flash accuracy is 0.617 without a definition, 0.911 with the source definition (56 repairs, 3 harms), 0.883 with the drafted gloss (52 repairs, 4 harms), and 0.544 with its own gloss (15 repairs, 28 harms).

\paragraph{Retrieval configurations.}
CUAD compares BM25, a dense BGE retriever with reranking, and hybrid BGE-M3 retrieval \citep{bm25,bge,bge-m3}. The configurations use eight retrieved passages. Reddit uses six passages. The CUAD full-glossary reference is capped at 30 entries. Risk selection uses a term maximum within each contract and a term mean within each Reddit community.

\begin{table}[t]\centering\small
\caption{In-sample hybrid-retrieval results on constructed questions. The risk aggregate includes the evaluated diagnostic item. The all-matched arm is unavailable for Reddit. The CUAD full glossary is capped at 30 entries. These arms are not matched-budget random or frequency policies.}
\label{tab:selection}
\begin{tabular}{@{}lrcccc@{}}\toprule
Workload & $n$ & Retrieval & Selected notes & All matched & Full glossary\\\midrule
CUAD & 1,200 & 0.821 & 0.871 & 0.882 & 0.876\\
Reddit / all & 1,391 & 0.866 & 0.925 & n/a & 0.907\\
Reddit / glossary & 851 & 0.839 & 0.941 & n/a & 0.907\\
Reddit / control & 540 & 0.907 & 0.898 & n/a & 0.907\\
\bottomrule\end{tabular}\end{table}

\paragraph{Overlap between scoring and evaluation questions.}
In both hybrid tests, the risk score for a term can include the score of the very question we evaluate. We remove that question's score and see which definition sets change. In CUAD, 990 of the 1,200 sets stay the same. Another 17 become empty. For those items, we already have a response with retrieval alone. For the remaining 193 changed sets, we have no response with the new set. The missing binary outcomes give a leave-one-question-out accuracy interval of $[0.733,0.893]$, compared with 0.821 for retrieval alone.

For Reddit, 1,295 of 1,391 sets stay the same. Another 33 become retrieval-only. For 63, we lack the response with the new set. The accuracy interval is $[0.874,0.919]$ over all questions, compared with 0.866 for retrieval alone. For the 851 glossary questions, it is $[0.859,0.933]$, compared with 0.839.

To get the bounds, we score every unobserved outcome as 0 at the lower end and 1 at the upper end. Reusing old retrieval or gated answers would give narrower scenarios, but not valid bounds. These intervals describe the observed workload and do not include clustered sampling uncertainty. Leaving out a question is also different from holding out an entire term or document. We have no random or frequency policy at the same budget in this workload. For Reddit, we also lack an all-matched arm.

The additional-passage arm is approximately cost-matched because it adds whole passages. The all-matched arm adds definitions for detected terms without risk gating. Definition lengths in Table~\ref{tab:supp-rag} exclude retrieval-only context and use the same regex counter within each workload.
\begin{table}[htbp]\centering\small
\caption{Full RAG accounting: selected-definition policy relative to retrieval only. Repairs and harms are paired counts. Added definition units are inferred by subtracting identical retrieval contexts. They include glossary formatting.}
\label{tab:supp-rag}
\begin{tabular}{@{}lrccc@{}}\toprule
Corpus / tier / subset & $n$ & Gain, pp [95\% CI] & Repair/harm & Selected/full units\\\midrule
CUAD / BM25 / all & 1200 & 4.17 [2.24, 6.12] & 94 / 44 & 135.8 / 529.2\\
CUAD / dense / all & 1200 & 5.33 [3.49, 7.35] & 102 / 38 & 79.1 / 529.2\\
CUAD / hybrid / all & 1200 & 5.00 [3.34, 6.73] & 93 / 33 & 126.6 / 529.2\\
Reddit / BM25 / all & 1391 & 6.69 [4.45, 9.04] & 112 / 19 & 39.4 / 2765.9\\
Reddit / BM25 / glossary & 851 & 10.34 [7.01, 13.83] & 97 / 9 & 46.5 / 3285.2\\
Reddit / BM25 / control & 540 & 0.93 [-0.89, 2.78] & 15 / 10 & 28.3 / 1947.5\\
Reddit / hybrid / all & 1391 & 5.90 [3.85, 7.77] & 106 / 24 & 38.5 / 2765.9\\
Reddit / hybrid / glossary & 851 & 10.22 [7.61, 13.30] & 92 / 5 & 46.9 / 3285.2\\
Reddit / hybrid / control & 540 & -0.93 [-2.86, 1.12] & 14 / 19 & 25.3 / 1947.5\\
\bottomrule\end{tabular}\end{table}

\paragraph{Inline selection.}
The 965 synthetic probes have 85 repairs and four harms. The 733 CUAD probes have 29 repairs and 16 harms. Full-injection net gains are 8.39 points with interval $[5.73,11.46]$ and 1.77 points with interval $[-0.44,3.91]$. An offline selection curve sums paired effects on selected probes and divides by the entire workload size. The random-allocation expectation is the selected fraction times the full-injection effect.

\paragraph{Other diagnostic comparisons.}
Self-report separates designed synthetic term classes more strongly than it predicts errors on real corpora. Frequency, word-list rarity, and contextual embedding shift have weaker reported associations with reader failures. Masked-span and span-likelihood estimators combine completion difficulty with local-meaning dependence. Sentence surprisal has little reported association with this risk target.

On the subset covered by the sampled Haiku proxy, it ranks errors less well than the local model. TypeSafe AI's Jev reports its own decision probabilities and confidence \citep{jev}. Negative native confidence gives a bare-error AUROC of 0.643 on 761 synthetic questions, 0.660 on 1,391 Reddit questions, and 0.636 on 671 CUAD questions. The synthetic set excludes the same two source-conflicted items as elsewhere.

With a drafted gloss, Jev's confidence gives AUROCs of 0.815, 0.790, and 0.868 on the three sets. Jev's error rate drops from 0.418 to 0.046 on the synthetic set. On Reddit it drops from 0.423 to 0.206. On CUAD it drops from 0.323 to 0.113. These measures rank Jev's own errors. They do not directly compare one reader with another using $R$. The supplementary inventory lists tests of model size, option order, and repeated question sets. The observed overlap between reader errors does not show that one set is a strict subset of another or that the error rate has a fixed lower limit.

Explicit Inverse Scaling redefinition items supply the local meaning in the prompt. Their compliance results concern following an available definition \citep{inverse-scaling,redefine-pitfalls}.

\paragraph{Decoding, scope, and carriers.}
We test Qwen2.5-14B-Instruct on 1,200 CUAD questions with hybrid retrieval. We compare greedy decoding, CAD, and AdaCAD \citep{cad,adacad}. With only the retrieved passages, CAD reaches 0.719 accuracy and greedy decoding reaches 0.747. The difference is $-2.8$ points, with a contract-cluster interval of $[-4.4,-1.0]$. Adding two usage examples from the same contract gives a difference of $-2.3$ points $[-4.0,-0.7]$. When we add definitions selected by risk, greedy decoding reaches 0.818 and CAD reaches 0.816. The difference is $-0.2$ points $[-1.7,+1.3]$. These comparisons cover one reader in the tested settings. They do not establish the cause of the errors or the performance of other decoding methods.

Plain, scoped, and negatively framed glossaries produce different rates of abstention on ordinary unlisted terms and over-application to usual-meaning questions. Scope wording reduces some of these effects. The carrier study compares full-context text, retrieval-scoped text, LoRA, and a GRACE-style editor \citep{lora,grace}. Outcomes of individual configurations do not establish a general limitation of either method family.

\section{Public human-written tasks}
\label{app:boundaries}
\begin{table}[htbp]\centering\small
\caption{Public-task interventions. Metric changes are in raw F1/accuracy units, with paired cluster-bootstrap intervals. Reddit labels refer to post-term items.}
\begin{tabular}{@{}lr lcc@{}}\toprule
Task & $n$ & Arm & Score & Difference [95\% CI]\\\midrule
CUAD F1 & 1661 & rag\_full & 0.5270 & +0.0029 [-0.0037, +0.0100]\\
CUAD F1 & 1661 & rag\_gated & 0.5249 & +0.0008 [-0.0051, +0.0073]\\
ContractNLI acc. & 3060 & note\_gated & 0.8173 & +0.0059 [-0.0020, +0.0147]\\
ContractNLI acc. & 3060 & note\_all & 0.8222 & +0.0108 [+0.0026, +0.0193]\\
Reddit / GPT & 178 & note\_gated & 0.5169 & +0.0169 [-0.0382, +0.0566]\\
Reddit / GPT & 178 & note\_all & 0.5281 & +0.0281 [-0.0715, +0.0909]\\
Reddit / DS & 178 & note\_gated & 0.6124 & +0.0056 [-0.0335, +0.0473]\\
Reddit / DS & 178 & note\_all & 0.5955 & -0.0112 [-0.0592, +0.0373]\\
\bottomrule\end{tabular}\end{table}
The CUAD extraction test uses 1,661 matched questions. Adding definitions changes token F1 by at most 0.3 points. ContractNLI \citep{contractnli} uses 3,060 hypotheses. We cluster by document. Adding all matched notes raises accuracy by 1.1 points, with an interval above zero. Adding notes selected by the gate raises it by 0.6 points. The public tests use different items from the diagnostic set. We did not measure how many benchmark items or natural requests depend on a definition boundary.

The Reddit comparison has 178 post-term items from 176 posts. People wrote the original posts and responses. The pipeline drafted the multiple-choice questions we evaluate. The TechQA \citep{techqa} check uses 105 drafted questions on 20 screened terms. It does not cover the full human-authored support task.

\section{Definition selection and source-text sensitivity}
\label{app:new}
\paragraph{CUAD development allocation.}
We draft the 63 evaluation questions separately from the A diagnostics. This is still a small, constructed task with forced choices. With every definition provided, GPT-5.6-terra gets 61 answers right, up from 51. DeepSeek V4.1 Flash gets 59 right, also up from 51. At the fixed 50\% question budget, mean $\Delta p$, matched-first, mean $R$, and frequency yield 55, 56, 57, and 59 correct GPT-5.6-terra answers, respectively. The corresponding DeepSeek V4.1 Flash counts are 56, 55, 55, and 56. No single $\Delta p$ policy consistently beats the simple alternatives across readers and budgets. We analyze this batch separately from the Reddit library experiment.

\paragraph{Shared-library split and definition versions.}
The Reddit calculation uses 215 glossary terms in 35 communities. Questions of a term are split by odd and even index into diagnostic A and evaluation B, so B is a sibling question rather than an independent future request. Each half usually has one or two questions per term. Frequency comes from the historical corpus. The library-policy comparison uses the exact community entry for A scoring, word cost, and B reader intervention. A separate drafted-gloss arm uses a different text for both scoring and intervention. Every drafted gloss differs from the corresponding community entry.

The proxy scores 429 A questions and the reader outcomes cover 422 B questions. The table below gives AUROC for predicting a wrong-to-correct transition conditional on a bare error. The A direct row scores the same question. The A-to-B row uses the A-term mean to score the other half. Intervals resample communities.

\begin{table}[htbp]\centering\small
\caption{Reddit conditional repair with the version of the definition held fixed between proxy scoring and reader intervention. These historical questions were model drafted and are not independent future requests.}
\label{tab:gold-source-repair}
\begin{tabular}{@{}p{2.8cm}p{4.3cm}p{4.3cm}@{}}\toprule
Definition text & A direct: bare failures / repairs, AUROC [95\% CI] & A to B: bare failures / repairs, AUROC [95\% CI]\\\midrule
Drafted gloss & 144 / 85, 0.833 [0.773, 0.882] & 147 / 85, 0.811 [0.732, 0.893]\\
Community source entry & 144 / 116, 0.665 [0.539, 0.788] & 147 / 126, 0.600 [0.475, 0.717]\\\bottomrule
\end{tabular}\end{table}

The community entry fixes more bare errors on B than the drafted gloss does. It also makes fewer correct answers wrong: four rather than 25. Only 21 bare errors remain after adding the community entry. This small group increases uncertainty in the ranking estimate. It does not by itself explain the point estimate of 0.600. Each AUROC refers to a particular definition text and its repair outcomes. The B question writer saw the community entry, and some keys lack independent review. The comparison is not a randomized test of definition quality.

\paragraph{Configured reader and answer quality.}
We set the budget to 25\% of the words in the source-entry library. Frequency per cost selects 111 definitions. Frequency times $\Delta p$ per cost selects 89. The selections differ on 60 terms. GPT-5.6-terra answers 118 B questions on those terms, first without a definition and then with the source entry. We record usage and response IDs for all 236 responses.

The terms unique to the frequency policy account for 549 definition words. Those unique to the $\Delta p$ policy account for 550. The two groups yield 26 and 11 repairs, respectively. The frequency group harms one answer. The $\Delta p$ group harms none. With the historical frequency weights, the $\Delta p$ policy produces 4.53 fewer correct outcomes per 100 eligible occurrences. The community-cluster interval is $[-9.45,-0.28]$ points for these fixed libraries. It does not include uncertainty about source validity or future use.

An outcome-blind DeepSeek V4.1 Flash source review accepts 83 of the 118 B questions under a fixed rule requiring one supported answer and no answer leakage. Fifteen differing terms have no accepted question. Bounding each missing term's mean repair effect by $[-1,1]$ gives a total difference interval of $[-9.21,+6.14]$ points. Model acceptance is not expert gold, and a fixed-hash spot review finds some remaining underspecified questions. The available evidence does not establish which library is better for future requests.

\section{Models and reproducibility}
\label{app:models}
The GPT readers are GPT-5.6-terra and GPT-6-astra, with API IDs \texttt{gpt-5.6-terra} and \texttt{gpt-6-astra}. The operator confirms that the synthetic DeepSeek calls before September 10, 2026 used V4 Flash. The later Reddit and DeFi calls used V4.1 Flash. DeepSeek documents \texttt{deepseek-flash} as the API ID for V4.1. After the September 10 release, the older \texttt{deepseek-v4-flash} ID routes to V4.1 \citep{deepseek-v41-release}. The later source screen also uses V4.1. We did not establish an official knowledge cutoff for either DeepSeek version.

GPT-5.6-terra answers the 1,500 CUAD validation questions. GPT-6-astra answers the additional synthetic, Reddit, and DeFi questions. The table lists the proxy checkpoints.

\begin{table}[h]
\centering\small
\caption{Proxy model checkpoints. Each entry is the first 12 characters of the Hugging Face commit hash that identifies the exact weights we used. The full 40-character hashes are in the accompanying model records.}
\begin{tabular}{@{}ll@{}}\toprule
Model & Hugging Face commit (first 12 characters)\\\midrule
Qwen2.5-0.5B & \texttt{060db6499f32}\\
Qwen2.5-1.5B & \texttt{8faed761d45a}\\
Qwen2.5-3B & \texttt{3aab1f1954e9}\\
Qwen2.5-7B & \texttt{d14972939875}\\
Qwen2.5-14B & \texttt{97e1e76335b7}\\
Qwen2.5-14B-Instruct & \texttt{cf98f3b3bbb4}\\
Llama-3.1-8B & \texttt{d04e592bb4f6}\\\bottomrule
\end{tabular}
\end{table}

Readers use greedy-style decoding at temperature zero where supported. For uncertainty, we draw ten samples per item at temperature 1.0. The proxy scores answer letters from local logits. RAG costs count context-length units rather than API billing tokens. Sonnet term means are reported by term. After excluding source conflicts, GPT-6 has prior-gloss responses for 440 of 456 synthetic items. The bare and local-definition conditions cover all 456.

Development logs record request settings, source and question hashes, model fields, token usage, response status, and outcomes by policy. Code and exports omit credentials. Deployment costs should separate building and maintaining the index from supplying notes. The index may be reused across requests.

\section{Exploratory analyses}\label{app:exploratory}
We include these exploratory analyses for completeness. They were inconclusive.

\paragraph{Clause alignment.}
Of 5,659 scored CUAD occurrences, 644 fall inside expert-labeled clause spans and 5,015 outside. The proxy probability change exceeds 0.3 for 0.1134 of items inside and 0.1163 outside. Clause labels describe legal topics rather than monetary or decision severity. Comparisons by category are exploratory. Categories may overlap.

\paragraph{Unrepaired errors.}
Of the 157 errors that remain after adding definitions, the model labels 53 as cases where the prior interpretation prevails. Other labels are question defects (39), missing relations (35), insufficient definitions (15), and a parameter issue (one). Fourteen have no label. The diagnostic and selection tests do not establish whether these labels describe actual causes.

\section{Supplementary experiment inventory}
The supplementary file \texttt{experiment\_inventory.tsv} lists every experiment, its evaluation unit, source hashes, and stored results.

\end{document}